\documentclass[a4paper]{article}

\usepackage{INTERSPEECH2020}
\usepackage{multirow}
\title{
Pardon the Interruption: An Analysis of Gender and Turn-Taking in U.S. Supreme Court Oral Arguments
}
\name{Haley Lepp$^1$, Gina-Anne Levow$^2$}
\address{
  $^1$Educational Testing Service, $^2$University of Washington,}
\email{$^1$hlepp@ets.org, $^2$levow@uw.edu}

\begin{document}

\maketitle
\begin{abstract}This study presents a corpus of turn changes between speakers in U.S. Supreme Court oral arguments. Each turn change is labeled on a spectrum of “cooperative" to “competitive" by a human annotator with legal experience in the United States. We analyze the relationship between speech features, the nature of exchanges, and the gender and legal role of the speakers. Finally, we demonstrate that the models can be used to predict the label of an exchange with moderate success. The automatic classification of the nature of exchanges indicates that future studies of turn-taking in oral arguments can rely on larger, unlabeled corpora.\footnote{This work was completed while the first author was a graduate student at the University of Washington.}
\end{abstract}

\section{Introduction}
The Supreme Court plays a key role in defining, identifying, and rooting out gender discrimination by hearing the cases that will determine the way gender rights are evaluated across the United States. However, there are few checks on the presence of gender bias within the court itself. This study offers a novel corpus of annotated speech from Supreme Court oral arguments and proposes a framework to analyze gender biases in turn changes.

For decades, scientists have argued that women are interrupted more than men in professional settings, indicating that this speech act could be an indicator of gender bias. The \textit{New York Times} has described ``being interrupted, talked over, shut down or penalized for speaking out" as ``nearly a universal experience for women when they are outnumbered by men" \cite{Chira17}.

Interruptions correlating with gender within Supreme Court oral arguments have occurred consistently over time, and are not necessarily due to political polarization or the personalities of justices \cite{Jacobi17}. However, in conversational turn-taking, an interruption is not inherently a negative act. As demonstrated by Tannen \cite{Tannen94} in her research on gender and language, interruptions cannot be defined categorically as acts of rudeness or dominance. Interruptions can be part of regular discourse depending on the context of a conversation, and are especially common among speakers of certain social groups in the United States.

Furthermore, the term ``interruption” is not a clear-cut linguistic term. Interruptions have variously been described as an overlap in speech between two speakers \cite{Yang03}, possibly including backchannels \cite{Laskowski10},  a ``power type” event to “wrest the discourse from the speaker”\cite{Goldberg90}, a ``topic change attempt”\cite{Goldberg90}, an event to ``bolster the interruptee’s positive face”\cite{Goldberg90}, or a syntactically incomplete turn \cite{wichmann10}. 

To address this, we annotate a corpus of turn changes as audio segments on a spectrum of cooperative to competitive.\footnote{The corpus of turn changes and annotations is available for public use at https://github.com/hlepp/pardontheinterruption.} To demonstrate the utility of the corpus, we extract speech features and show that classifiers can automatically predict the human labels of the turn changes with relative success. 

\section{Audio and transcription retrieval}

\subsection{Original audio and transcript files}
The transcripts and audio recordings of all U.S. Supreme Court oral arguments since October 2006 are publicly available online. The transcripts, written by court stenographers, are formatted like the script of a play, with the name of the speaker followed by the transcribed speech. The transcripts include disfluencies and speech that ends mid-sentence or word \cite{SCOTUS}.

The transcriptions do not include the time at which each statement is said, so we retrieve time stamps of turns or sentences (whichever are shorter) from \textit{The Oyez Project} \cite{Oyez}.

\subsection{Turn extraction}

We define a turn change as an event in which one speaker stops speaking and a second speaker starts speaking according to the transcript. 
For example:

\begin{quote}Hannah S. Jurss:  And so we're certainly asking for this Court's -- \\
John G. Roberts, Jr.: But I'm not faulting them for that.\cite{MitchellvWisconsin}\end{quote}

Using the start and end time-stamp of each speaker’s turn, we segment each argument into short audio clips around turn changes. Multiple studies have demonstrated that listeners can perceive significant social and emotional information from a short slice of an audio, despite not knowing the greater context of a conversation~\cite{Ambady06, Ambady93}. The brevity of clips also ensures that the annotators, who are busy professionals, do not lose interest in the task. Also, because this study aims to find patterns in speech without regard to the subject matter of the case, limiting the content which an annotator can listen to can help avoid annotator bias.  

The default length of a segment is six seconds long: two seconds before the end-label of the first speaker and four seconds after the start label of the next speaker. If the turn of the first speaker is less than two seconds long, then we use the start of the first speaker’s turn as the start of the turn change, instead of a full two seconds of audio. If the turn of the second speaker is less than four seconds long, then we use the end of that speaker’s turn as the end of the turn change.

We manually check each segment and remove those in which at least one speaker is inaudible (probably due to the stenographer hearing something not picked up by the microphone), turns that are listed as separate in the transcripts but are the same person with a pause, and turns that are scripted, such as “Mr. Chief Justice, and may it please the court.” We trim recordings by no more than one second if another adjacent turn change occurs that makes it unclear what an annotator might be labeling. We extend recordings by no more than one second if the change becomes more clear with extension; this is usually due to timestamp rounding cutting off a very short turn.

We update speaker names if the ordering is wrong or names were incorrect. For example, if a number of turns occur in quick succession or there are two or more speakers talking at the same time, we change the label so that the first and second speakers heard are the first and second speakers listed. Exchanges with fully overlapped speech are checked to ensure the order of speakers aligns with human perception.  

\subsection{Turn change corpus}
The corpus includes 711 turn changes from four oral arguments: Kahler v. Kansas \cite{KahlervKansas}, Mitchell v. Wisconsin \cite{MitchellvWisconsin}, Virginia House of Delegates v. Bethune-Hill \cite{HoDvVA}, and Washington State Dept. of Licensing v. Cougar Den Inc. \cite{WAvCoug}. A typical case is heard by nine justices (three female, six male) and two or three attorneys (the corpus includes seven female and five male). Each of the selected trials occurred in 2018 or 2019, covers a unique topic, and includes at least one female arguing before the court.\footnote{The latter qualification narrows the selection considerably: in 2018, only 15\% of the people who argued before the court were women \cite{Strawbridge19, Walsh18}.},\footnote{Gender information was gathered from public profiles of speakers. An expansion of the corpus should ensure such characteristics align with speakers’ self-identities.}

The number of turns per attorney in the corpus ranges from 27 to 128. For justices, each of whom appear in every oral argument, the number of turns per individual per trial ranges from 10 to 42; with one exception: Justice Clarence Thomas does not speak in any of the four arguments. Among justices, Justice Sonia Sotomayor and Justice Stephen Breyer are most represented, with over 130 turns each. 

\begin{table}[ht]
\caption{Information about Annotated Segments}
\centering
\begin{tabular}{| l | l |}
\hline
\textbf{Corpus Component} & \textbf{Number} \\
\hline
Male Participants & 11, with Justice Thomas\\
\hline
Female Participants &
9 \\
\hline
Justice to Attorney exchanges &
338 \\
\hline
Attorney to Justice exchanges &
351 \\
\hline
Justice to Justice exchanges &
22 \\
\hline
Attorney to Attorney exchanges &
0 \\
\hline
Female to Female exchanges &
127 \\
\hline
Male to Male exchanges &
269 \\
\hline
Female to Male exchanges &
165 \\
\hline
Male to Female exchanges &
150 \\
\hline
\end{tabular}
\label{tab:segmentdata}
\end{table}

\section{Corpus annotation}
\subsection{Motivation for a survey of legal professionals}
The rules of conversational speech within a courtroom setting are not the same as those in informal conversational speech; power-relationships, formal rules and procedures, and field-specific argument strategies are among the many factors that influence the ways that speakers interact within an oral argument. 
In the annotation process, all annotators are required to be U.S.-based, and to identify as an attorney, judge, legal scholar, or law student in their second year or above.

\subsection{Survey design}
We design a brief, easy-to-complete, anonymous online survey for legal professionals. The survey is built on the JavaScript library JSPsych \cite{jspsych}. We instruct participants to categorize the short clips on a spectrum of “cooperative” to “competitive.”  Before beginning annotation, the participants are given descriptions of each category:
\begin{quote}
By cooperative, we mean that to your ears, the first speaker expects a turn change and gives the floor to the second speaker. The second speaker might leave space for the first speaker to finish their turn. Or, the second speaker might talk at the same time as the first speaker, providing short spurts of feedback, for example saying ``mhmm" or ``yes".	

By competitive, we mean that to your ears, the second speaker competes with the first speaker for the chance to speak. You, the first speaker, or listeners might perceive this as a disruption to the previous speaker's speech. The second speaker may cause the first speaker to stop speaking, or talk over the first speaker to compete to be heard.
\end{quote}

as well as trial exercises and example audio clips that could be classified into each category.
The descriptions demonstrate a division between the two categories of turn changes, but make clear that the participant should use their training and experience to evaluate turn changes with nuance.

After this short training, the annotators proceed to the tasks. Each task includes the prompt “How competitive or cooperative do you perceive this exchange to be?”, which emphasizes to the participant that the annotation should be solely from their perception. Below this prompt is an audio element which the user can control and a slider showing a spectrum from Competitive to Cooperative with Likert-style category labels. The participants are instructed to leave the slider in the middle if the category of the clip is unclear. The speaker is given up to 26 more segments to listen to (a number selected to keep the entire annotation exercise under five minutes). If the speaker leaves the survey before completion, all results are still saved. 

\subsection{Survey participants}
Each segment is annotated twice by 2 of 77 unique annotators. Each speaker annotated between 1 and 26 segments, with a mean per person of 18, and a standard deviation of 6.2.

As the way listeners perceive speech differs depending on many factors, we ask participants to optionally share demographic data to demonstrate that the age, gender, ethnic, political, and linguistic diversity of listeners is relatively representative of the diversity of the United States. This information is included in the corpus.

\subsection{Annotation distribution}
\begin{figure}[h!]
\begin{center}
  \includegraphics[width=.45\textwidth]{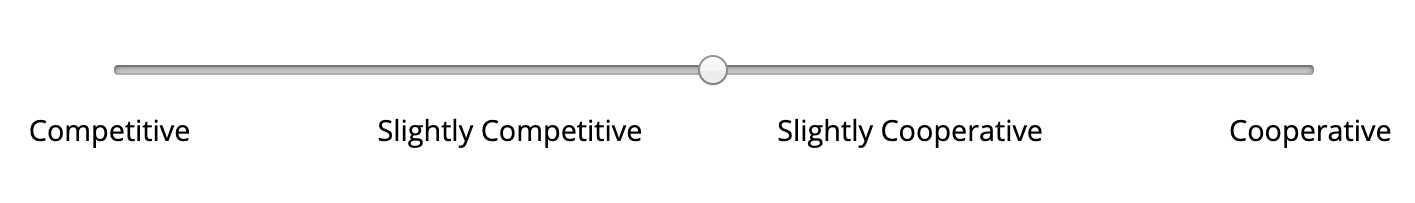}
    \caption{Slider given to annotators.}
        \label{fig:AnnotationSpectrum}
\end{center}
\end{figure}

Annotators score each audio segment on a visual spectrum, as seen in Figure~\ref{fig:AnnotationSpectrum}. The location on which an annotator places the slider is codified with a score between 0 and 100, in which 0 represents the most competitive turn change, and 100 is the most cooperative turn change. 

\begin{figure}[h!]
\begin{center}
   \includegraphics[width=0.49\textwidth]{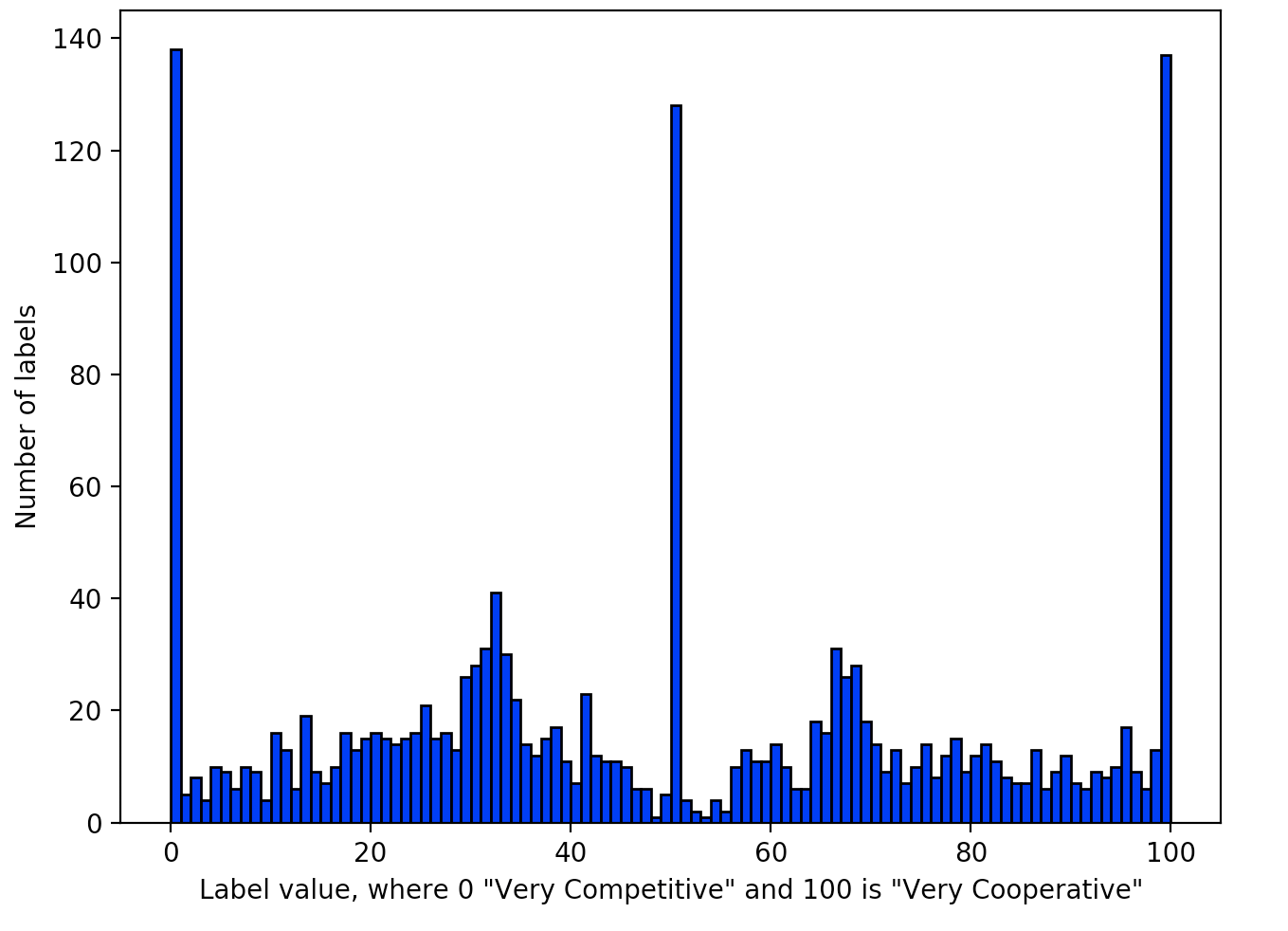}
    \caption{The distribution of labels in the annotated corpus.}
        \label{fig:Distribution}
\end{center}
\end{figure}

The distribution of labels in Figure~\ref{fig:Distribution} reflects the layout of the web interface. The highest peaks are at either end of the spectrum and directly in the middle. This phenomenon indicates that annotators move the slider all the way to one end when an audio clip clearly sounds competitive or cooperative. The annotators leave the slider in place if the audio does not clearly fall into a category. There are also middling peaks around where the survey interface has labels of “slightly competitive” and “slightly cooperative.” These peaks show that annotators make use of the Likert-style guidelines, despite having the ability to drop the button anywhere on the slider. 
\subsection{Annotation assessment}
We evaluate the annotations under the assumption that if the audio files receive similar annotations by different people, then the annotation process can be considered reproducible \cite{Artstein17}. Inter-annotator agreement on the raw labels as well as labels categorized into five equally-spaced bins are shown in Table~\ref{fig:agreement} and indicate moderate agreement on this task.

\begin{table}[ht]
\label{tab:AnnotatorAgreement}
\caption{Annotator agreement.}\label{fig:agreement}
\centering
  \begin{tabular}{| l | l | l |}
\hline
 & Raw & Five Bins \\
\hline
Spearman's $\rho$ & 0.556 & 0.547\\
\hline
Weighted Cohen's $\kappa$ & 0.553 & 0.542 \\
\hline
\end{tabular}
\end{table}
Qualitative analysis shows that high disagreement appears to arise from conflicting cues. For example, in some clips the speakers talk over each other but the content of their speech is polite: phrases like “If I may, your honor” \cite{MitchellvWisconsin} and “Sorry, sorry your honor” \cite{WAvCoug} might sway the annotator to mark an otherwise competitive interaction as cooperative.

\section{Data analysis}
To investigate the relationship between gender and competitive turn-taking, we explore the distribution of turn change scores with respect to the gender of the first speaker in the turn. For comparison, we also consider the distribution of these turn change scores with respect to the role of the speaker (i.e. justice or attorney).

The mean score of an exchange when a woman is the first speaker is more competitive (45.0 with a standard deviation of 28.1) than when a man is the first speaker (49.7 with a standard deviation of 26.7). Alternatively, the mean label for a turn in which a woman is the second speaker is slightly more cooperative (51.5) than when a man is (45.0). More extreme is the difference between roles: the mean label for an attorney first speaker is 36.3, while a justice first speaker has a much more cooperative mean of 59.0. This is predictable considering the power differential and a culture of deference; attorneys would avoid speaking competitively to a justice, while justices would be much more likely to speak to an attorney competitively. There are no instances of attorney-to-attorney speech.

\begin{figure}[h!]
\begin{center}
  \includegraphics[width=.48\textwidth]{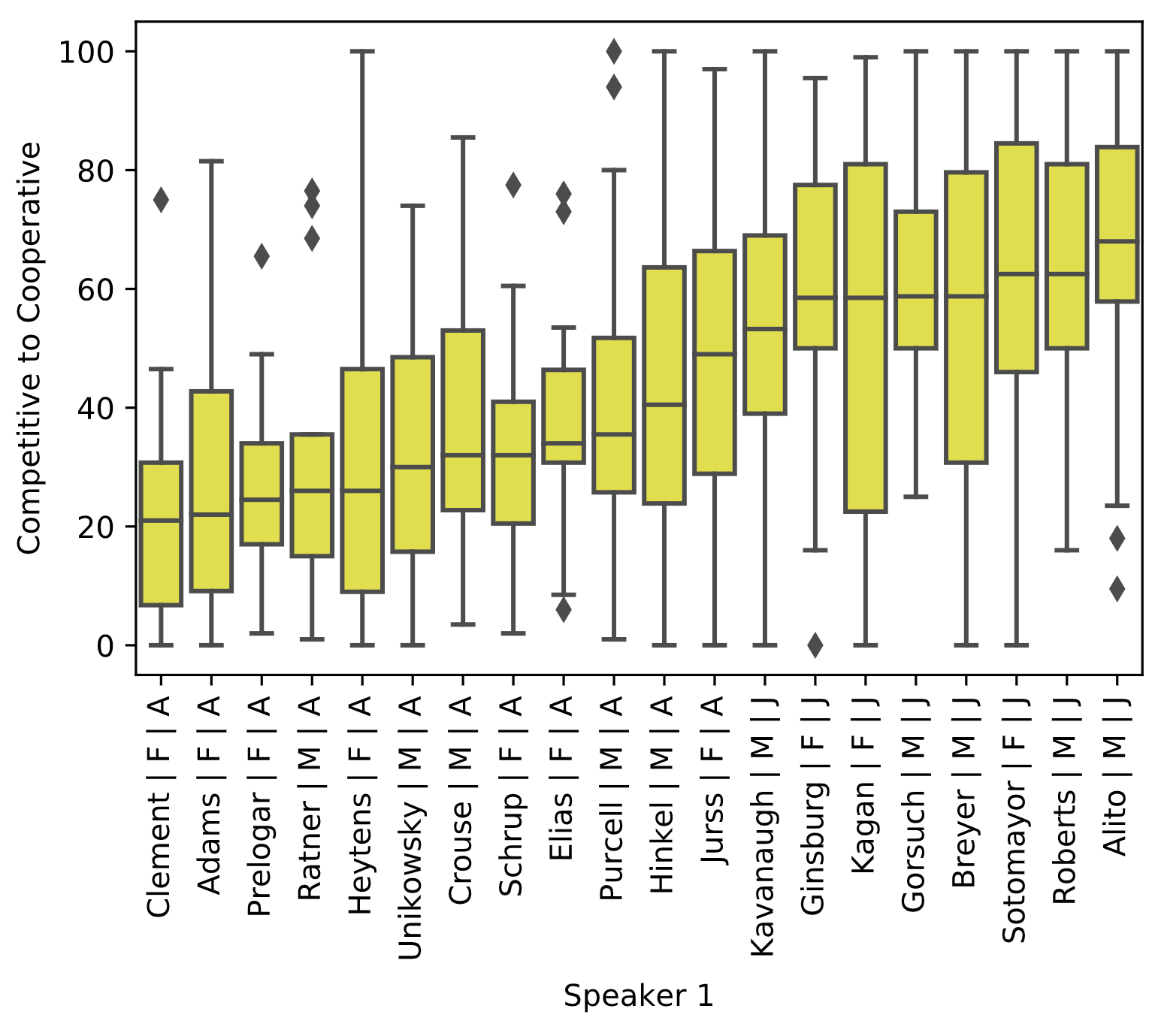}
    \caption{Distribution of labels for the first speaker in a turn.}
        \label{fig:spokento}
\end{center}
\end{figure}

Figure \ref{fig:spokento} shows the distribution of labels for each speaker in the corpus who is the first speaker in a turn. The distribution illuminates the severity with which role aligns with turn change type. The eight speakers who have the highest means, or are spoken to most cooperatively, are all justices; those with the lowest means are all attorneys.

Based on the non-parametric Kruskal-Wallis test, we confirm a significant effect of speaker gender ($p < 0.03$) and of speaker role ($p = 7.012e^{-29}$) on competitiveness score. A Wilcoxon Ranked-Sum test also shows this difference, with comparably low $p$-values.

\section{Turn Classification Experiments}
We also investigate the automatic classification 
of turn changes based on acoustic and speaker 
cues.  Effective classification could enable analysis
of the relationship between gender and turn change
at scale.

\subsection{Feature Extraction}
 Using \texttt{openSMILE}, we extract time-aggregated features for each speaker in each audio segment in the corpus. We use two feature sets: the eGeMaPS collection of 88 psychologically-informed features, and the Speech Prosody collection of 36 pitch and loudness related features \cite{OpenSmile, Eyben15}. We select eGeMaPs because of its demonstrated success in emotion recognition studies, and we select the Speech Prosody set because patterns in pitch and amplitude have been shown linguistically to differentiate  between cooperative and competitive turn-taking \cite{Gorisch12, Truong13, Yang03, wichmann10}. In each feature set, we also include the gender and role of the first and second speakers. We do not normalize pitch features for speaker gender.\footnote{We find that normalization of features by speaker gender harms predictive results, and leave it to future research to explore this phenomenon more.}

\subsection{Classification}
We divide the labeled corpus into two subsets: 80\% of the corpus into a training set and 20\% into an evaluation set. Each set has comparable gender and role distribution across turn-type; for example, 21\% of the turns in the full corpus, training set, and evaluation set are male-to-female, and 49\% of all turns in each set are attorney-to-justice. 

To group the labels into classes, we take the mean of a segment's two raw labels given by annotators in a 0 to 100 scale, then categorize each segment into one of three quantile-based classes based on that mean: the most competitive, the most cooperative, and the middling exchanges. 

We measure the effectiveness of Random Forest (RF) and Support Vector Machine classifiers (SVC with RBF kernel) in predicting whether an audio segment falls into these competitive and cooperative classes. We use the SciKit Learn Laboratory Toolkit version 2.0 (SKLL)\footnote{https://skll.readthedocs.io/en/latest/index.html}, with features scaled by standard deviation and centered around a mean, and a micro-averaged $F_{1}$ score as a grid search objective \cite{scikit-learn}. The training data is divided randomly into subsets for grid search for hyperparameter optimization. 

We report two baselines. The first predicts a competitive turn in every instance in which the transcription of the first speaker's speech ends in a dash ("-"), indicating syntactic incompleteness, and a cooperative turn when there is no dash. Second, we show the micro $F_{1}$ score were the target class predicted for all instances.
\subsection{Results}
The classifiers are most successful at predicting the competitive turns, with the highest micro $F_{1}$ resulting from the Speech Prosody feature set. This may be due to the fact that high pitch and loudness are defining elements of a competitive turn change, while there may be more variation in what could define a cooperative turn change. The results without adding gender and role as features, as well as normalizing prosodic features by gender, are generally lower, but within 0.1 of listed respective scores. The added performance due to these features could be due to the fact that gender and role do correlate with the class of segments. 

\begin{table}[ht]
\caption{Micro-$F_{1}$ score for baseline predictions of classes}\label{Resultslast} 
\centering
  \begin{tabular}{| l | l | l |} 
\hline
&\textbf{Dash} &\textbf{Target Class}  \\
\hline
\textbf{Competitive} & 0.328 (dash) & 0.338\\
\hline
\textbf{Cooperative}& 0.494 (no dash) & 0.331 \\
\hline
\end{tabular}
\end{table}
\begin{table}[ht]
\caption{Micro-$F_{1}$ score for SVC and RF predictions of classes}\label{Resultslast} 
\centering
 \begin{tabular}{| l | l | l | l |} 
 \hline
&\textbf{Model} & \textbf{eGeMaPS} & \textbf{Prosody} \\\hline
\multirow{ 2}{*}{\textbf{Competitive}} & 
SVC &0.636&\textbf{0.687}\\ 
& RF &0.617&0.640\\
\hline
\multirow{ 2}{*}{\textbf{Cooperative}} &
SVC &0.551&0.561\\ 
 & RF &0.593&\textbf{0.611}\\\hline
\end{tabular}
\end{table}

\section{Conclusion}
This study introduces a corpus of segments of speech from U.S. Supreme Court oral arguments that include a turn change between speakers. The segments, annotated by legal practitioners for competitiveness and cooperativeness, provide insight in the ways that justices and attorneys speak with one another in this unique speech setting. We find that as the first person in an exchange, female speakers and  attorneys are spoken to more competitively than are male speakers and  justices. We also find that female speakers and attorneys speak more cooperatively as the second person in an exchange than do male speakers and justices. We demonstrate that classifiers trained only on phonetic and acoustic features extracted from the audio segments can achieve a level of predictive accuracy above multiple baselines. 

In-depth studies of gender bias and inequality are critical to the oversight of an institution as influential as the Supreme Court. While the models presented in this study analyze linguistic trends in relation to gender, the labeled corpus could be integrated with other demographic or content-related information to provide a fine-grained analysis of intersectional fields. There is demand in the social sciences for even broader analysis; within the first few months of 2020, several cross-cutting studies have criticized increasing bias in the Supreme Court and federal appeals courts, especially in regards to poverty and race \cite{Ruiz20, Cohen20}. With improved predictive models, a larger set of turn changes across all Supreme Court oral argument recordings and possibly other court recordings could provide fodder for future statistical social science studies of speech trends in the U.S. judicial system.
\section{Acknowledgments}
 We are grateful to the scholars who supported this cross-disciplinary study: Richard Wright, Keelan Evanini, Vikram Ramanarayanan, Victoria Zayats, and the board, reviewers, and participants in Widening NLP at ACL 2019. We also thank the legal professionals who annotated data and helped test the annotation survey. Finally, we express our appreciation for the public servants and activists who spend their daily lives fighting for equality and fairness in the U.S. Judicial System.

\bibliographystyle{IEEEtran}


\begin{thebibliography}{10}
\providecommand{\url}[1]{#1}
\csname url@samestyle\endcsname
\providecommand{\newblock}{\relax}
\providecommand{\bibinfo}[2]{#2}
\providecommand{\BIBentrySTDinterwordspacing}{\spaceskip=0pt\relax}
\providecommand{\BIBentryALTinterwordstretchfactor}{4}
\providecommand{\BIBentryALTinterwordspacing}{\spaceskip=\fontdimen2\font plus
\BIBentryALTinterwordstretchfactor\fontdimen3\font minus
  \fontdimen4\font\relax}
\providecommand{\BIBforeignlanguage}[2]{{%
\expandafter\ifx\csname l@#1\endcsname\relax
\typeout{** WARNING: IEEEtran.bst: No hyphenation pattern has been}%
\typeout{** loaded for the language `#1'. Using the pattern for}%
\typeout{** the default language instead.}%
\else
\language=\csname l@#1\endcsname
\fi
#2}}
\providecommand{\BIBdecl}{\relax}
\BIBdecl

\bibitem{Chira17}
S.~Chira, ``{The Universal Phenomenon of Men Interrupting Women},'' \emph{The
  New York Times}, 06 2017.

\bibitem{Jacobi17}
T.~Jacobi and D.~Schweers, ``\BIBforeignlanguage{English (US)}{{Justice,
  Interrupted: The Effect of Gender, Ideology and Seniority at Supreme Court
  Oral Arguments}},'' \emph{\BIBforeignlanguage{English (US)}{Virginia Law
  Review}}, vol. 103, no.~7, pp. 1379--1496, 11 2017.

\bibitem{Tannen94}
D.~Tannen, \emph{Gender and Discourse}.\hskip 1em plus 0.5em minus 0.4em\relax
  New York: Oxford University Press, 1994.

\bibitem{Yang03}
L.~Yang, \emph{Current and New Directions in Discourse and Dialogue. Text,
  Speech and Language Technology}.\hskip 1em plus 0.5em minus 0.4em\relax
  Springer, 2003, vol.~22, ch. Visualizing Spoken Discourse.

\bibitem{Laskowski10}
K.~Laskowski, ``Modeling norms of turn-taking in multi-party conversation,'' in
  \emph{Proceedings of the 48th Annual Meeting of the Association for
  Computational Linguistics}.\hskip 1em plus 0.5em minus 0.4em\relax
  Association for Computational Linguistics, 2010, p. 999–1008.

\bibitem{Goldberg90}
J.~A. Goldberg, ``Interrupting the discourse on interruptions: An analysis in
  terms of relationally neutral, power and rapport-oriented acts,''
  \emph{Journal of Pragmatics}, vol.~14, pp. 883--903, 12 1990.

\bibitem{wichmann10}
\BIBentryALTinterwordspacing
A.~Wichmann and J.~Caspers, ``Melodic cues to turn-taking in {E}nglish:
  Evidence from perception,'' in \emph{{Proceedings of the Second SIGdial
  Workshop on Discourse and Dialogue}}, 2001. [Online]. Available:
  \url{https://www.aclweb.org/anthology/W01-1625}
\BIBentrySTDinterwordspacing

\bibitem{SCOTUS}
{The Supreme Court of the United States}, ``Transcripts and {R}ecordings of
  {O}ral {A}rguments,'' 2019.

\bibitem{Oyez}
{The Oyez Project}, ``About {O}yez,'' https://www.oyez.org/about, 2019.

\bibitem{MitchellvWisconsin}
{Mitchell v. Wisconsin}, \emph{No. 18-6210}.\hskip 1em plus 0.5em minus
  0.4em\relax {United States Supreme Court}, Jan. 21, 2019.

\bibitem{Ambady06}
N.~Ambady, M.~A. Krabbenhoft, and D.~Hogan, ``{The 30-Sec Sale: Using
  Thin-Slice Judgments to Evaluate Sales Effectiveness},'' \emph{Journal of
  Consumer Psychology}, vol.~16, pp. 4--13, 2006.

\bibitem{Ambady93}
N.~Ambady and R.~Rosenthal, ``{Half a minute: Predicting teacher evaluations
  from thin slices of nonverbal behavior and physical attractiveness},''
  \emph{Journal of Personality and Social Psychology}, vol.~64, p. 431–441,
  1993.

\bibitem{KahlervKansas}
{Kahler v. Kansas}, \emph{No. 18-6135}.\hskip 1em plus 0.5em minus 0.4em\relax
  {United States Supreme Court}, Oct. 7, 2019.

\bibitem{HoDvVA}
{Virginia House of Delegates v. Bethune-Hill}, \emph{No. 18-281}.\hskip 1em
  plus 0.5em minus 0.4em\relax {United States Supreme Court}, Mar. 18, 2019.

\bibitem{WAvCoug}
{Washington State Dept. of Licensing v. Cougar Den Inc.}, \emph{No.
  16-1498}.\hskip 1em plus 0.5em minus 0.4em\relax {United States Supreme
  Court}, Oct. 30, 2018.

\bibitem{Strawbridge19}
K.~S. Robinson and J.~S. Rubin, ``{Women Argue Only a Fraction of Supreme Court
  Cases},'' Jan. 30, 2019.

\bibitem{Walsh18}
M.~Walsh, ``{Number of Women Arguing Before the Supreme Court has Fallen off
  Steeply},'' \emph{American Bar Association Journal}, Aug. 1, 2018.

\bibitem{jspsych}
J.~R. de~Leeuw, ``{jsPsych: A JavaScript library for creating behavioral
  experiments in a web browser},'' \emph{Behavior Research Methods}, vol.~47,
  pp. 1--12, 2015.

\bibitem{Artstein17}
R.~Artstein, ``{Inter-annotator Agreement},'' in \emph{Handbook of Linguistic
  Annotation}, P.~J. Ide~N., Ed.\hskip 1em plus 0.5em minus 0.4em\relax
  Springer, Dordrecht, 2017, ch.~11, pp. 297--313.

\bibitem{OpenSmile}
\BIBentryALTinterwordspacing
F.~Eyben and B.~Schuller, ``{OpenSMILE:): The Munich Open-Source Large-Scale
  Multimedia Feature Extractor},'' \emph{SIGMultimedia Rec.}, vol.~6, no.~4, p.
  4–13, Jan. 2015. [Online]. Available:
  \url{https://doi.org/10.1145/2729095.2729097}
\BIBentrySTDinterwordspacing

\bibitem{Eyben15}
F.~{Eyben}, K.~R. {Scherer}, B.~W. {Schuller}, J.~{Sundberg}, E.~{André},
  C.~{Busso}, L.~Y. {Devillers}, J.~{Epps}, P.~{Laukka}, S.~S. {Narayanan}, and
  K.~P. {Truong}, ``The {G}eneva {M}inimalistic {A}coustic {P}arameter {S}et
  ({GeMAPS}) for voice research and affective computing,'' \emph{IEEE
  Transactions on Affective Computing}, vol.~7, no.~2, pp. 190--202, 2016.

\bibitem{Gorisch12}
J.~Gorisch, B.~Wells, and G.~Brown, ``{Pitch Contour Matching and Interactional
  Alignment Across Turns: An Acoustic Investigation},'' \emph{Language and
  {S}peech}, vol.~55, pp. 57--76, 03 2012.

\bibitem{Truong13}
K.~Truong, ``{Classification of Cooperative and Competitive Overlaps in Speech
  Using Cues from the Context, Overlapper, and Overlappee},'' \emph{Proceedings
  of the Annual Conference of the International Speech Communication
  Association (INTERSPEECH)}, pp. 1404--1408, 2013.

\bibitem{scikit-learn}
F.~Pedregosa, G.~Varoquaux, A.~Gramfort, V.~Michel, B.~Thirion, O.~Grisel,
  M.~Blondel, P.~Prettenhofer, R.~Weiss, V.~Dubourg, J.~Vanderplas, A.~Passos,
  D.~Cournapeau, M.~Brucher, M.~Perrot, and E.~Duchesnay, ``{Scikit-learn:
  Machine Learning in Python},'' \emph{Journal of Machine Learning Research},
  vol.~12, pp. 2825--2830, 2011.

\bibitem{Ruiz20}
R.~R. Ruiz, R.~Gebeloff, S.~Eder, and B.~Protess, ``{A Conservative Agenda
  Unleashed on the Federal Courts},'' \emph{The New York Times}, 03 2020.

\bibitem{Cohen20}
A.~Cohen, \emph{Supreme Inequality: The Supreme Court's Fifty-Year Battle For a
  More Unjust America}.\hskip 1em plus 0.5em minus 0.4em\relax New York:
  Penguin Random House, 2020.

\end{thebibliography}


\end{document}